\documentclass[sigconf,nonacm]{acmart}

\AtBeginDocument{%
  }

\setcopyright{none}
\acmConference{}{}{}
\acmPrice{}
\acmISBN{}
\copyrightyear{}
\acmYear{}
\acmDOI{}
\usepackage{microtype}
\usepackage{graphicx}
\usepackage{subfigure}
\usepackage{booktabs} 

\usepackage{hyperref}

\usepackage{amsmath}

\usepackage{amssymb}
\DeclareMathOperator*{\Exp}{\mathbb{E}}
\usepackage[capitalize,noabbrev]{cleveref}

\title{Avoiding Generative Model Writer's Block With Embedding Nudging }
\author{Ali Zand}
\email{zand@google.com}
\affiliation{%
  \institution{Google }
  \city{}
  \state{}
  \country{}
}

\author{Milad Nasr}
\email{srxzr@google.com}
\affiliation{%
  \institution{Google Deepmind}
  \city{}
  \state{}
  \country{}
}

\date{September 2023}

\begin{document}

\begin{abstract}

Generative image models, since introduction, have become a global phenomenon. From new arts becoming possible to new vectors of abuse, many new capabilities have become available. One of the challenging issues with generative models is controlling the generation process specially to prevent specific generations classes or instances . There are several reasons why one may want to control the output of generative models, ranging from privacy and safety concerns to application limitations or user preferences

To address memorization and privacy challenges, there has been considerable research dedicated to filtering prompts or filtering the outputs of these models. What all these solutions have in common is that at the end of the day they stop the model from producing anything, hence limiting the usability of the model. In this paper, we propose a method for addressing this usability issue by making it possible to steer away from unwanted concepts (when detected in model's output) and still generating outputs. In particular we focus on the latent diffusion image generative models and how one can prevent them to generate particular images while generating similar images with limited overhead.

We focus on mitigating issues like image memorization, demonstrating our technique's effectiveness through qualitative and quantitative evaluations. Our method successfully prevents the generation of memorized training images while maintaining comparable image quality and relevance to the unmodified model.

\end{abstract}
\maketitle
\section{Introduction}

Following recent advances (Song et al.,~\cite{song2019generative}; Ramesh et al.,~\cite{ramesh2021zero}; Ho et al.,~\cite{ho2020denoising}), diffusion models have emerged as leading technology for generating images, videos, and audio. This advancement has paved the way for numerous exciting applications, such as image editing and super-resolution, time-series forecasting, and many others.

With diffusion models, as with any other generative model trained on real data, privacy and memorization are real concerns. Researchers have suggested three different ways to address memorization: controlling the input, controlling the output, and controlling the training process. Some researchers suggest removing sensitive input from the training dataset to address privacy concerns, real-world implementation proves challenging. For instance, in image generation, determining which images belong to which entities and whether to remove them entirely is not always clear-cut. This complexity intensifies with memorization concerns.
Other researchers have suggested controlling/filtering the output of these models, but ensuring the output adheres to specific constraints without introducing usability problems remains a challenge. This often leads to the primary solution for filtering unwanted output being a preventative filter rather than a corrective one. There are several reasons why we might want to control the output of generative models, ranging from privacy and memorization concerns to application limitations or user preferences.

A third group of researchers have suggested alternative training techniques such as differential privacy offer a theoretical solution by defining sensitivity broadly, but they often come at the cost of significantly reduced model performance or rely on large theoretical bounds without a concrete level of protection. Furthermore, these methods are data agnostic which means that they do not consider if the input is images or email or any other type, therefore it is complicated to make them work for specific use cases.

In this work, we take a different approach to controlling the output of generative models. Instead of focusing on the training dataset, we focus on controlling the model's output at inference time. In particular, we focus on the image generative models. This approach allows us to address controlling the properties of the image generation model concerns without directly modifying the training pipeline which might be more complex and challenging.

We focus on mitigating image memorization and demonstrate the effectiveness of our technique on the Stable Diffusion 1.4 model. Through both qualitative and quantitative evaluations, we show that our method successfully prevents the generation of memorized training images while maintaining comparable image quality and relevance to the unmodified model. Our approach opens up new possibilities for fine-grained control over generative models, enhancing their applicability in privacy-sensitive contexts and other scenarios where specific outputs need to be avoided.

\section{Background}
\paragraph{Image Generative Models} Image generative models have been a subject of research for many years, with notable contributions \cite{hinton2006fast, goodfellow2020generative, salakhutdinov2010efficient, xie2016theory, kingma2013auto, vincent2010stacked, li2015generative, uria2016neural} (see~\cite[Chapter 20]{goodfellow2016deep}). The introduction of Generative Adversarial Networks (GANs) by~\cite{goodfellow2020generative}  marked a significant breakthrough, enabling the generation of high-quality images at scale as demonstrated by Brock et al.~\cite{brock2018large} and Karras et al.~\cite{karras2019style}.

However, in recent years, diffusion models, initially proposed by Sohl-Dickstein et al.~\cite{sohl2015deep}, have surpassed GANs in terms of performance. These models have achieved state-of-the-art results on academic benchmarks (Dhariwal and Nichol, ~\cite{dhariwal2021diffusion}) and serve as the foundation for popular image generators like Stable Diffusion (Rombach et al.~\cite{rombach2022high}), DALL-E 2 (Ramesh et al.~\cite{ramesh2021zero,ramesh2022hierarchical}), Runway (Rombach et al.,~\cite{rombach2022high}), Midjourney, and Imagen (Saharia et al.,~\cite{saharia2022photorealistic}).

Denoising Diffusion Probabilistic Models (DDPMs), introduced by Ho et al.~\cite{ho2020denoising}, operate on a simple principle: they are image denoisers. During training, a clean image $x$ is perturbed by adding Gaussian noise at a sampled timestep $t$, resulting in a noised image $x'$. The diffusion model $f_{\theta}$ is trained to predict the added noise, effectively recovering the original image $x$. This is achieved by minimizing the objective function:
\begin{equation}
\frac{1}{N} \sum_i \Exp_{\substack{x \sim P_{data} \\ t \sim \mathcal{U}[0, 1] \\ \omega \sim \mathcal{N}(0, I)}}
 [ \mathcal{L}(x_i, t, \epsilon; f_{\theta}) ], 
\end{equation}

where
\begin{equation*}
\mathcal{L}(x_i, t, \epsilon; f_{\theta}) = \lVert \epsilon - f_{\theta}(\sqrt{a_t} x_i + \sqrt{1 - a_t} \epsilon, t) \rVert_2^2.
\end{equation*}

Despite their simple training objective, diffusion models can generate high-quality images by iteratively applying the model to a random noise image $z_T$. This process gradually removes noise, resulting in a final image $z_0$ that resembles a natural image. Some diffusion models are further conditioned on class labels or text embeddings to generate specific types of images.

Class-conditional diffusion models take a class label (e.g., "dog" or "cat") along with the noised image to generate an image of that class. Text-conditional models take a text embedding of a more general prompt (e.g., "a photograph of a horse on the moon") using a pre-trained language encoder like CLIP (Radford et al.,~\cite{radford2021learning}).

\paragraph{Controlled Generations}  While several papers have explored control theoretic aspects of diffusion models (Liu et al.~\cite{}, Zhang et al.~\cite{}), their focus has been on enhancing training or sampling processes directly through control theory. Our work, however, diverges by concentrating on inference-level approaches and maintaining the core diffusion process unchanged.

\section{Methodology}
Diffusion-based models have demonstrated the capability to transfer artistic styles by guiding the latent representation towards desired embeddings. We propose an approach that leverages this mechanism to not only attract the latent representation towards desirable styles but also to repel it away from undesirable image representations.

\begin{figure}
    \centering
    \includegraphics[scale=0.6]{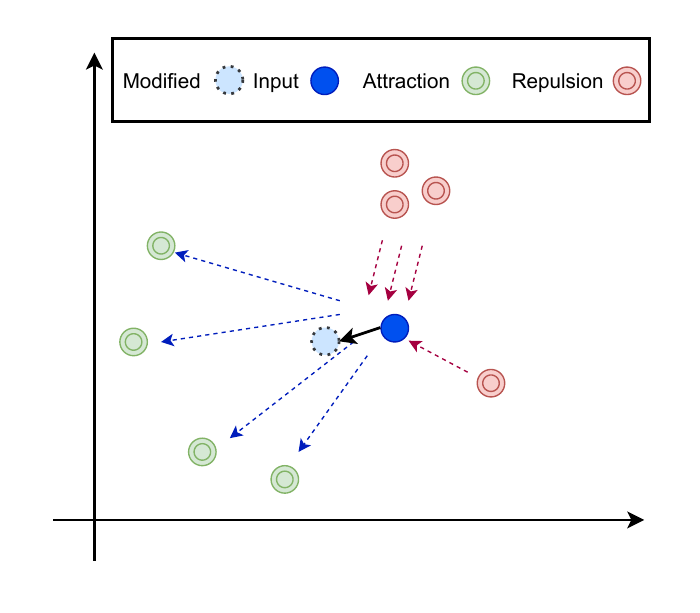}
    \caption{Overview of the methodology. We optimize the latent representation of the diffusion model to avoid generating unwanted outputs.}
    \label{fig:overview}
\end{figure}

\begin{figure*}[h!]
    \centering
    \includegraphics[scale=0.6]{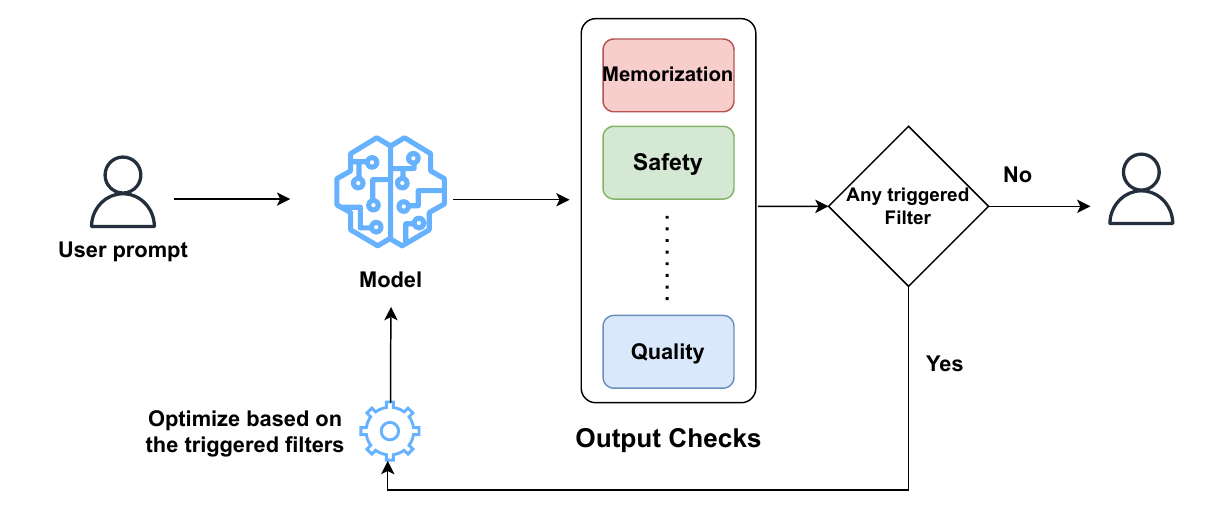}
    \caption{Our adaptive image generation system uses a series of filters to assess output quality. If filters are triggered, the system dynamically selects points to adjust the generation and improve it. This approach only adds computational cost when filters identify potential issues, leaving unproblematic prompts unmodified.}
    \label{fig:design}
\end{figure*}

Our method generalizes this concept, enabling simultaneous attraction to preferred styles and repulsion from undesired instances. Pushing is achieved by modifying the latent space through optimization to increase  the cosine similarity between the latent representation and an array of preferred embeddings representing desired styles. Pulling, conversely, is achieved by  optimizing the latent space to reduce  the cosine similarity between the latent representation and a set of embeddings representing undesired images. Figure~\ref{fig:overview} summarizes the approach. Formally we optimize the following objective function to refine the latent input to the image generative model.

\begin{equation}
   \min_{L}   -\alpha \frac{\sum_{D \in I_D}{D \times L}}{|I_D|} + \beta \frac{\sum_{U \in I_U}{U \times L}}{|I_U|}  + |L-L_0| \label{eq:main}
\end{equation}

Where $L_0$ represents the original latent variable, while $I_D$ and $I_U$ denote the sets of latent variables for desired and undesired images, respectively. The refined latent variable $L$ is obtained through a process of pulling towards $I_D$ and pushing away from $I_U$, controlled by coefficients $\alpha$ and $\beta$. These coefficients dictate the degree of deviation from the original prompt and should be carefully tuned for optimal performance in various applications.

\section{Experiments}

\paragraph{Setup} We focus our experiments on the Stable Diffusion~\cite{}. Given several of the previous works focus on Stable diffusion 1.4 version we also focus on that version to show effect of our approach. We modified the original stable diffusion inference pipeline. In particular we implemented a gradient descent based optimizer to minimize the objective in Equation~\ref{eq:main}. In this paper we mostly focus on using the approach to prevent memorization, the benefit of using this approach is that we can easily test if the the approach is successful or not. In particular, we follow the memorization definition used by Carlini et al~\cite{}. 

\begin{figure*}[t]
\centering
\hspace{-1cm}
\begin{tabular}{ll}
     \begin{tabular}{r}\,\,\,\,Original:\end{tabular} \vspace{1.3em}&  \includegraphics[width=1.85cm]{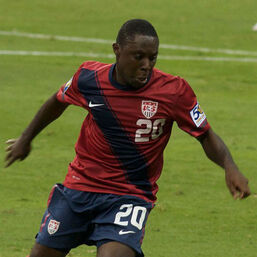}\includegraphics[width=1.85cm]{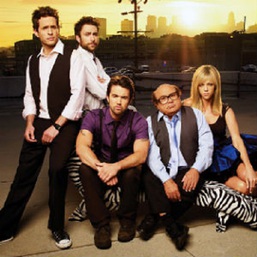}\includegraphics[width=1.85cm]{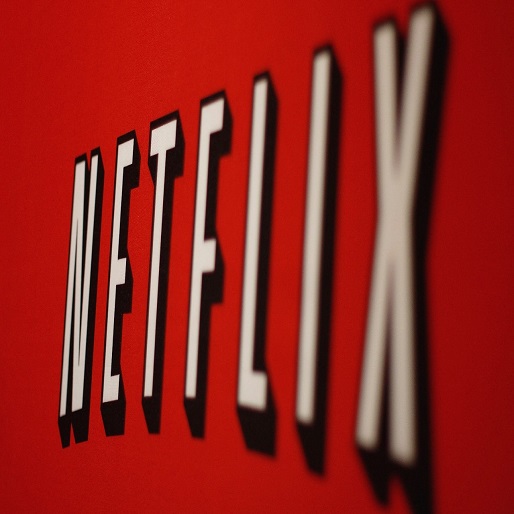}\includegraphics[width=1.85cm]{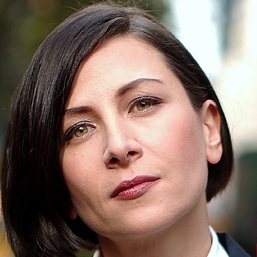}\includegraphics[width=1.85cm]{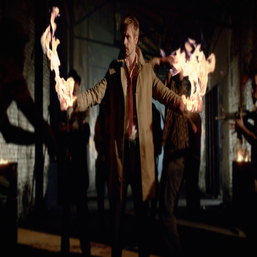}\includegraphics[width=1.85cm]{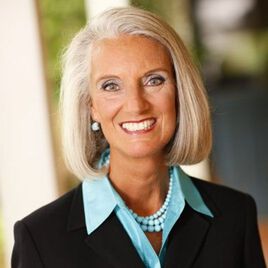}\includegraphics[width=1.85cm]{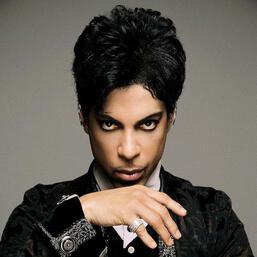}\\
     \begin{tabular}{r}Initial Generated:\end{tabular} \vspace{1.3em}& \includegraphics[width=1.85cm]{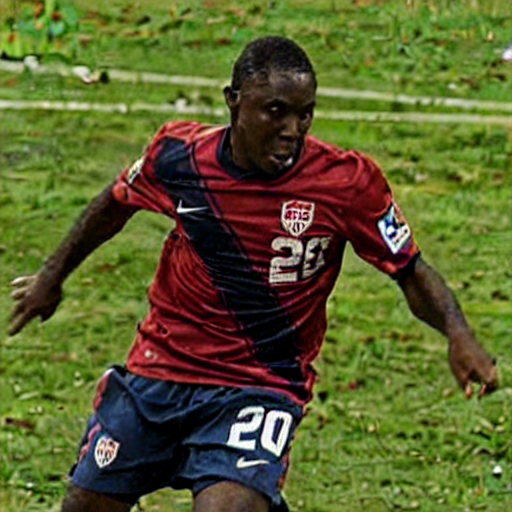}\includegraphics[width=1.85cm]{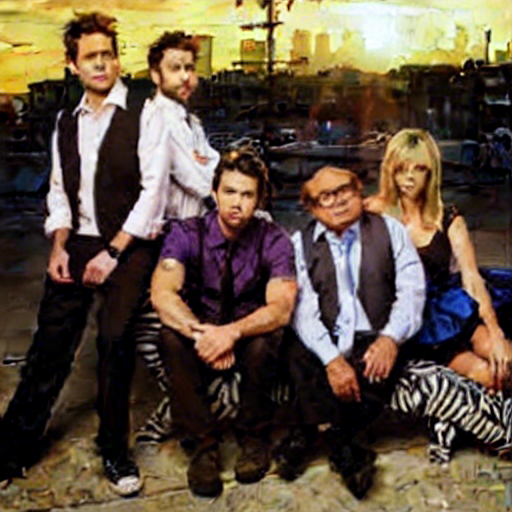}\includegraphics[width=1.85cm]{sample_images/198848_original.jpeg}\includegraphics[width=1.85cm]{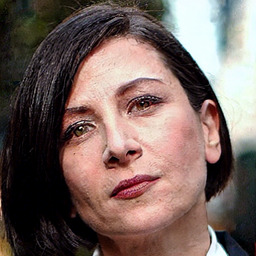}\includegraphics[width=1.85cm]{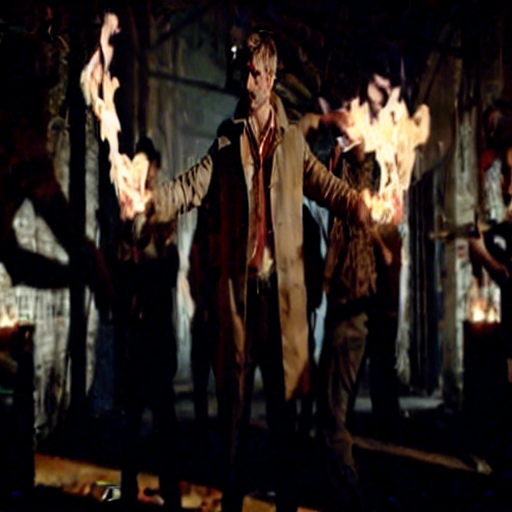}\includegraphics[width=1.85cm]{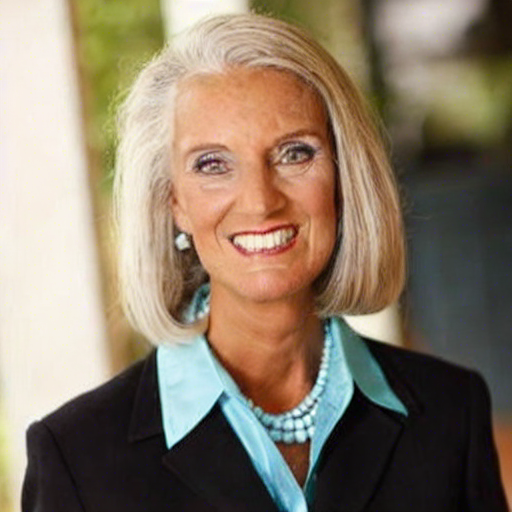}\includegraphics[width=1.85cm]{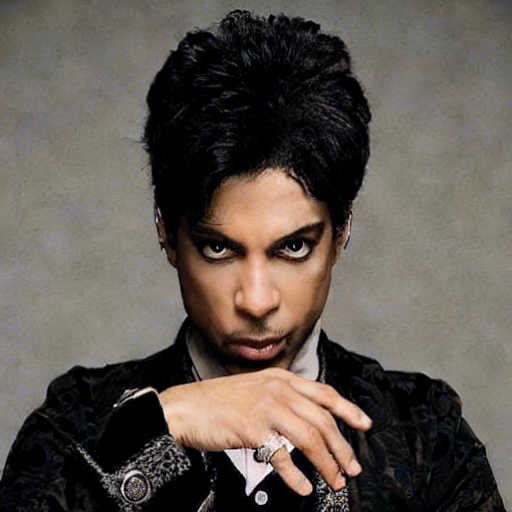} \\
     \begin{tabular}{r}{Output}:\end{tabular} \vspace{1.3em}& \includegraphics[width=1.85cm]{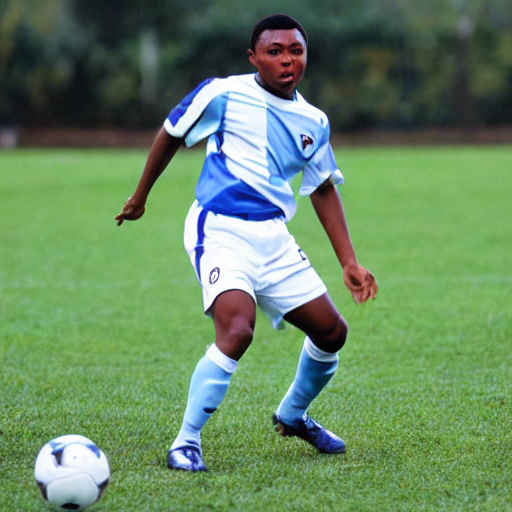}\includegraphics[width=1.85cm]{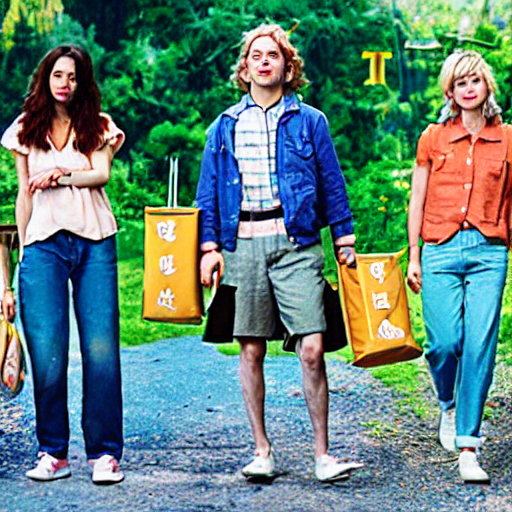}\includegraphics[width=1.85cm]{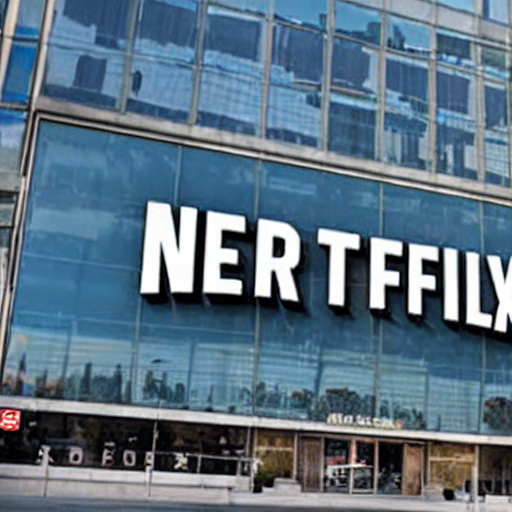}\includegraphics[width=1.85cm]{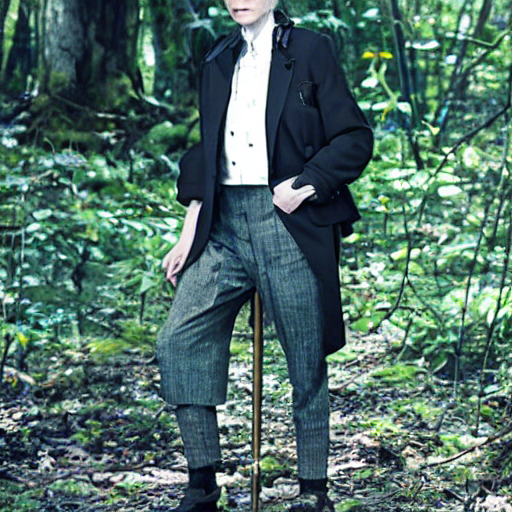}\includegraphics[width=1.85cm]{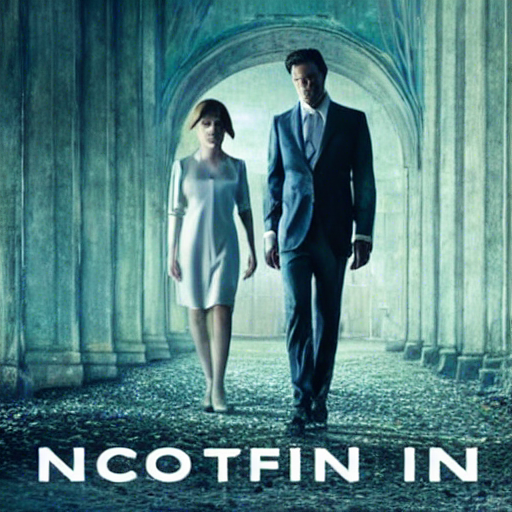}\includegraphics[width=1.85cm]{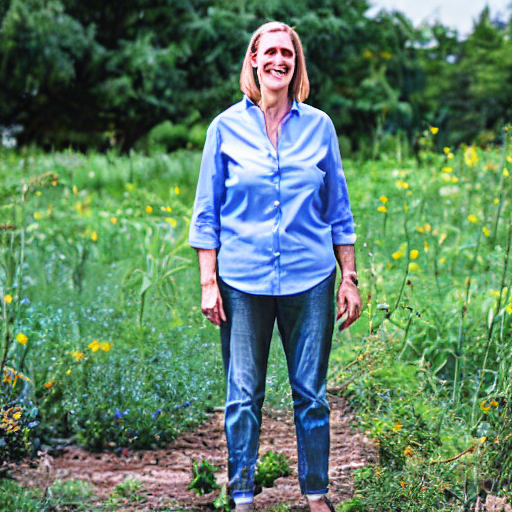}\includegraphics[width=1.85cm]{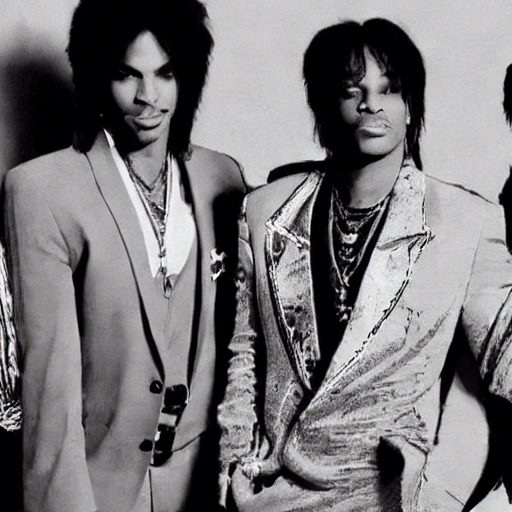}
\end{tabular}
\caption{Our approach prevents memorization of training images, generating similar yet distinct images from the same captions. This maintains image relevance while mitigating overfitting concerns.}
    \label{fig:demo}
\end{figure*}

\paragraph{Design}  Our method relies on having a set of points to push against and pull toward in order to guide image generation process. One way to achieve this is by using predefined points, but this may not always result in optimal model performance. Instead, we propose an adaptive system that begins by taking user input and generating an initial output. This output is then passed through a series of filters that assess the quality of the generation from various angles. If any filters are triggered, we dynamically select points to push against and pull toward to address the identified issues, followed by another round of generation. Figure~\ref{fig:design} summarizes our design approach. Given the assumption that most prompts are unproblematic, our approach leaves them unmodified and only incurs additional computational cost for user inputs that trigger the filters. Please note that this paper is not about the filters themselves, as they can be of a variety of types (e.g., images too similar to training material, NSFW filters, etc), but is about what to do when a filter is triggered and has stopped the generation of a specific image.

\subsection{Training Data Prevention}
While image quality can be subjective and debated, one clear issue is the memorization of training images by models. This is undesirable for various reasons and has a clear definition, making it a suitable focus for our study. Building on Carlini et al.~\cite{carlini2023extracting}'s findings of image memorization in Stable Diffusion, we evaluate our system's ability to mitigate this problem. Specifically, we test our system's performance on known prompts that generate memorized images, pushing away from these images in the generation process. Using image captions from the training dataset as prompts, we generate modified images and assess whether these altered versions are still present in the training data.

\begin{figure}
    \centering
    \includegraphics[scale=0.45]{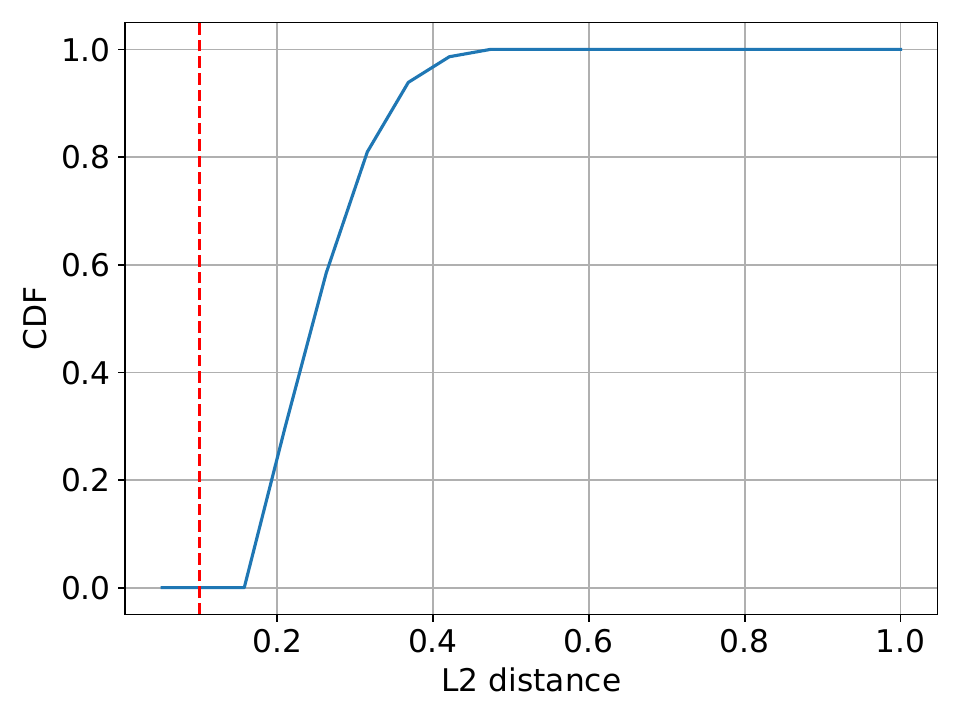}
    \caption{Distrubiton of the l2 distance between the generated images and the target images}
    \label{fig:l2dist}
\end{figure}

Figure ~\ref{fig:demo} demonstrates results where inputting an image's caption into the original model reproduces training images, as shown in Carlini et al.~\cite{carlini2023extracting}. This highlights the issue of image memorization and the potential for models to overfit on training data. By contrast, our approach, when applied to the same captions, generates similar images without directly replicating the training examples. This suggests our method can effectively mitigate generating memorizated images while still maintaining the ability to generate relevant and diverse images.

Following the "tiled" l2 distance metric from Carlini et al.~\cite{carlini2023extracting}, we measure the maximum l2 distance between corresponding 128x128 tiles in pairs of images, ensuring a stricter measure of similarity. We use 10 different random seeds to push away from known memorized images when generating new ones. Figure ~\ref{fig:l2dist} shows the distribution of normalized l2 distances for these generated images. All distances are high, exceeding the memorization threshold (0.1) established used in Carlini et al~\cite{carlini2023extracting} therefore none of the generated images will be triggered the memorization filters.

\subsection{Image Quality Test}


\begin{figure}
    \centering
\includegraphics[scale=0.35]{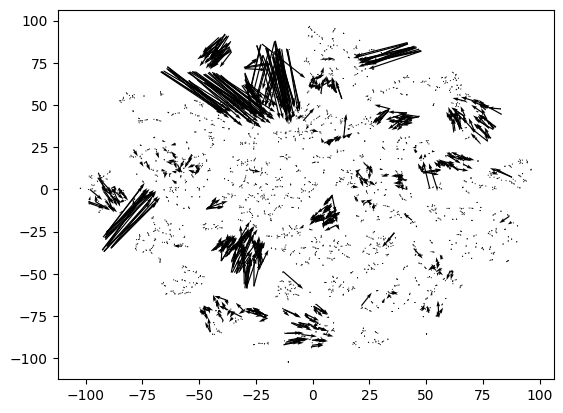}
    \caption{Arrows representing the changes from unmodified image generation to the modified versions using our approach in 2D projection of CLIP embeddings.}
    \label{fig:arrow}
\end{figure}
 In this  section we will study the effect of our approach on the relevancy and quality of the generations.

\paragraph{Embedding approach}
To demonstrate that images generated with our approach remain similar to those generated by the unmodified model, we compute the CLIP embeddings of both image sets and project them into a 2D space for visualization. Figure~\ref{fig:arrow} plots arrows from the unmodified generation to the modified version using our approach. This allows us to visually assess the magnitude and direction of changes induced by our method.

In most cases, our approach introduces minimal changes, preserving the relevancy and diversity of the original model's output. This is crucial as it ensures that our method doesn't significantly alter the intended image content or style. However, in some instances, larger differences are observed, but the modified images still remain within the same general region of the embedding space, indicating a degree of semantic similarity. This suggests that while our approach may introduce some variations, it does not fundamentally alter the core characteristics of the generated images.

\begin{figure}
    \centering
    \includegraphics[scale=0.5]{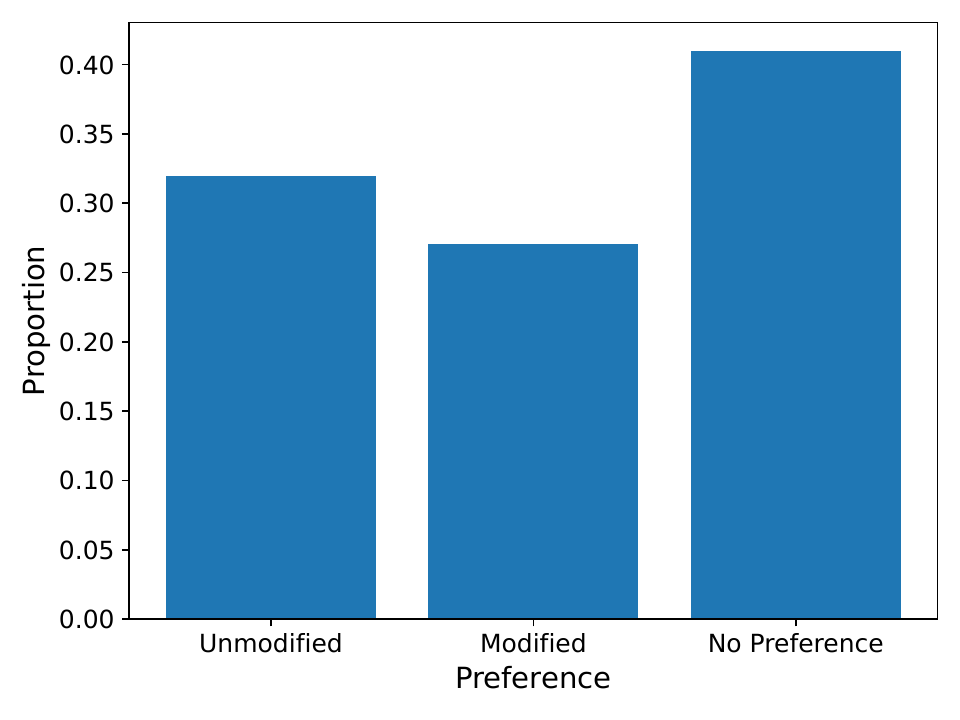}
    \caption{Results of our  user study with 200 participants, over $40\%$ showed no preference between images generated by our approach and the unmodified model.}
    \label{fig:preferences}
\end{figure}

\paragraph{Human Evaluations}

We conducted a user study with 200 participants, including both computer science scholars and Amazon Mechanical Turk workers, to assess the quality of images generated by our approach. Participants were shown pairs of images generated from the same prompt: one with the unmodified original model and one with our modified model that pushes away from memorized images. They were asked to indicate which image better represented the given prompt.

As shown in Figure~\ref{fig:preferences}, over $40\%$ of respondents expressed no preference between the two versions, while only $32\%$ preferred the original unmodified output. These results demonstrate that the quality and relevance of images generated by our approach are comparable to those of the unmodified version.

\section{Conclusions}
Our proposed method for controlling the output of image generative models, specifically latent diffusion models, presents a promising solution to the challenge of preventing unwanted generations while maintaining usability. By steering the generation process away from undesirable concepts through a combination of pulling to preferred styles and pushing from undesired instances, we were able to effectively mitigate issues like image memorization without significantly compromising the quality or relevance of the generated images.

Our experiments, focusing on Stable Diffusion 1.4, demonstrated the effectiveness of our approach in preventing the generation of memorized images from the training dataset, while still producing diverse and relevant outputs. The evaluation of image quality, both through embedding analysis and human evaluations, further confirmed that our method maintains comparable quality and relevance to the unmodified model. While our work primarily focused on preventing memorization, the flexibility of our approach allows for potential applications in various other areas where controlling the output of image generative models is desired. Future research can explore these applications and further refine the methodology for broader impact.
\bibliographystyle{plain}
\bibliography{main}
\end{document}